# Fast classification using sparse decision DAGs

Djalel Benbouzid[1]                                          DJALEL.BENBOUZID@GMAIL.COM
Róbert Busa-Fekete[1,2]                                              BUSAROBI@GMAIL.COM
Balázs Kégl[1]                                                  BALAZS.KEGL@GMAIL.COM

[1]LAL/LRI, University of Paris-Sud, CNRS, 91898 Orsay, France

[2]Research Group on Artificial Intelligence of the Hungarian Academy of Sciences and University of Szeged, Aradi vértanúk tere 1., H-6720 Szeged, Hungary

## Abstract

In this paper we propose an algorithm that builds sparse decision DAGs (directed acyclic graphs) from a list of base classifiers provided by an external learning method such as AdaBoost. The basic idea is to cast the DAG design task as a Markov decision process. Each instance can decide to use or to skip each base classifier, based on the current state of the classifier being built. The result is a sparse decision DAG where the base classifiers are selected in a data-dependent way. The method has a single hyperparameter with a clear semantics of controlling the accuracy/speed trade-off. The algorithm is competitive with state-of-the-art cascade detectors on three object-detection benchmarks, and it clearly outperforms them when there is a small number of base classifiers. Unlike cascades, it is also readily applicable for multi-class classification. Using the multi-class setup, we show on a benchmark Web page ranking data set that we can significantly improve the decision speed without harming the performance of the ranker.

## 1. Introduction

There are numerous applications where the computational requirements of classifying a test instance are as important as the performance of the classifier itself. Object detection in images (Viola & Jones, 2004) and web page ranking (Chapelle & Chang, 2011) are well-known examples. A more recent application domain with similar requirements is trigger design in high energy physics (Gligorov, 2011). Most of these applications come with another common feature: the negative class (usually called noise or background) sometimes has orders of magnitudes higher probability than the positive class. Besides the testing time constraints, this also makes training difficult: traditional classification-error-based measures are not adequate, and using prior class probabilities in constructing training samples leads to either enormous data sizes or little representativity of the positive class.

A common solution to these problems is to design *cascade classifiers* (Viola & Jones, 2004). A cascade classifier consists of *stages*. In each stage a binary classifier attempts to eliminate background instances by classifying them negatively. Positive classification in inner stages sends the instance to the next stage, so detection can only be made in the last stage. By using simple and fast classifiers in the first stages, "easy" background instances can be rejected fast, shortening the expected testing time. The cascade structure also allows us to use different training sets in different stages, having more difficult background samples in later stages.

Cascade classifiers, however, have many disadvantages in both the training and test phases. The training process requires a lot of hand-tuning of control parameters, and it is non-trivial how to handle the trade-off between the performance and the complexity of the cascade. Also, each individual stage needs to be trained with examples that have been classified positively by all the previous stages, which becomes difficult to satisfy in the later stages. Moreover, during test time, the cascade structure itself has several drawbacks. First, for a given stage, the margin information of a test example is lost and is not exploited in the subsequent stages. Second, all the positive instances have to pass through all the stages for a correct clas-sification. Finally, extending the cascade architecture to the multi-class case is non-trivial. For example in web page ranking, it is just as crucial to make a fast prediction on the relevance of a web page to a query as in object detection (Chapelle et al., 2011a), but unlike in object detection, human annotation often provides more than two relevance levels.

In this paper we propose a method intended to overcome these problems. In our setup we assume that we are given a sequence of low-complexity, possibly multiclass base classifiers (or features) sorted by importance or quality. Our strategy is to design a *controller* or a *decision maker* which decides which base classifiers should be evaluated for a given instance. The controller makes its decision sequentially based on the output of the base classifiers evaluated so far. It has three possibilities in each step: 1) it can decide to continue the classification by evaluating the next classifier, 2) skip a classifier by jumping over it, or 3) quit and use the current combined classifier. The goal of the controller is to achieve a good performance with as few base classifier evaluations as possible. This flexible setup can accommodate any performance evaluation metric and an arbitrary computational cost function. Designing the controller can be naturally cast into a Markov decision process (MDP) framework where the roles are the following: the policy is the controller, the index of the base classifier and the output of classifier constitute the states, the alternatives correspond to the actions, and the rewards are defined based on the target metric and the cost of evaluating the base classifiers.

Our approach has several advantages over cascades. First, we can eliminate stages. Similar to SOFTCASCADE (Bourdev & Brandt, 2005), the base classifiers do not have to be organized into a small number of stages before or while learning the cascade. Second, we can easily control the trade-off between the average number of *evaluated* base classifiers and the quality of the classification by combining these two competing goals into an appropriate reward. The form of the reward can also easily accommodate costsensitivity (Saberian & Vasconcelos, 2010) of the base classifiers although we will not investigate this issue here. The fact that some base classifiers can be skipped has an important consequence: the resulting classifier is *sparse*, moreover, the number and identities of base classifiers depend on the particular instances. Third, eliminating stages allows each instance to "decide" its own path within the list of base-classifiers. Theoretically, we could have as many different paths as training instances, but, a-posteriori, we observe clustering in the "path-space". Fourth, eliminating stages also greatly simplifies the design. Our algorithm is basically turn-key: it comes with an important design parameter (the trade-off coefficient between accuracy and speed) and a couple of technical hyperparameters of the MDP algorithm that can be kept constant across the benchmark problems we use. Finally, the multi-class extension of the technique is quite straightforward.

Allowing skipping is an important feature of the algorithm. The result of this design choice is that the structure of the learned classifier is not a cascade, but a more general *directed acyclic graph* or a *decision DAG*. In fact, the main reason for sticking to the cascade design is that it is easy to control with semi-manual heuristics. Once the construction is automatic, keeping the cascade architecture is no longer a necessary constraint. Allowing skipping is also a crucial difference compared to the approach of (Póczos et al., 2009) who also proposed to learn a cascade in an MDP setup. While their policy simply designs optimal thresholds in stages of a classical cascade, MDDAG outputs a classifier with a different structure. Our method can also be related to the sequential classifier design of (Dulac-Arnold et al., 2011). In their approach the action space is much larger: at any state the controler can decide to jump to *any* of the base classifiers, and so the action space grows with the number of base learners. Whereas this design choice makes feature selection more flexible, it also generates a harder learning problem for the MDP.

The paper is organized as follows. In Section 2 we describe the algorithm, then in Section 3 we present our experimental results. In Section 4 we discuss the algorithm and its connection with existing methods, and in Section 5 we draw some pertinent conclusions.

## 2. The MDDAG algorithm

We will assume that we are given a sequence of $N$ *base classifiers* $\mathcal{H} = (\mathbf{h}_1, \ldots, \mathbf{h}_N)$. Although in most cases cascades are built for binary classification, we will describe the method for the more general multiclass case, which means that $\mathbf{h}_j : \mathcal{X} \to \mathbb{R}^K$, where $\mathcal{X}$ is the input space and $K$ is the number of classes. The semantics of $\mathbf{h}$ is that, given an observation $\mathbf{x} \in \mathcal{X}$, it *votes for* class $\ell$ if its $\ell$th element $h_\ell(\mathbf{x})$ is positive, and votes against class $\ell$ if $h_\ell(\mathbf{x})$ is negative. The absolute value $|h_\ell(\mathbf{x})|$ can be interpreted as the confidence of the vote. This assumption is naturally satisfied by the output of ADABOOST.MH (Schapire & Singer, 1999), but in principle any algorithm that builds its final classifier as a linear combination of simpler functions can be used to provide $\mathcal{H}$. In the case of ADABOOST.MH or multi-class neural networks, the *final* (or *strong* or *averaged*) classifier defined by the full sequence $\mathcal{H}$ is $\mathbf{f}(\mathbf{x}) = \sum_{j=1}^{N} \mathbf{h}_j(\mathbf{x})$, and its prediction for the class index of $\mathbf{x}$ is $\widehat{\ell} = \arg\max_{\ell} f_{\ell}(\mathbf{x})$. In binary *detection*, $\mathbf{f}$ is usually used as a scoring function. The observation $\mathbf{x}$ is classified as positive if $f_1(\mathbf{x}) = -f_2(\mathbf{x}) > \theta$ and background otherwise. The threshold $\theta$ is a free parameter that can be tuned to achieve, for instance, a given false positive rate.

The goal of the MDDAG (Markov decision direct acyclic graph) algorithm is to build a *sparse* final classifier from $\mathcal{H}$ that does not use all the base classifiers, and which selects them in a way depending on the instance $\mathbf{x}$ to be classified. For a given observation $\mathbf{x}$, we process the base classifiers in their original order. For each base classifier $\mathbf{h}_j$, we choose from among three possible actions: 1) we EVALUATE $\mathbf{h}_j$ and continue, 2) we SKIP $\mathbf{h}_j$ and continue, or 3) we QUIT and return the classifier built so far. Let

$$b_j(\mathbf{x}) = 1 - \mathbb{I}\left\{a_j = \text{SKIP} \vee \exists j' < j : a_{j'} = \text{QUIT}\right\} \quad (1)$$

be the indicator that $\mathbf{h}_j$ is evaluated on $\mathbf{x}$, where $a_j \in \{\text{EVAL}, \text{SKIP}, \text{QUIT}\}$ is the action taken at step $j$ and the indicator function $\mathbb{I}\{A\}$ is 1 if its argument $A$ is true and 0 otherwise. Then the final classifier built by the procedure is

$$\mathbf{f}^{(N)}(\mathbf{x}) = \sum_{j=1}^{N} b_j(\mathbf{x})\mathbf{h}_j(\mathbf{x}). \quad (2)$$

The decision on action $a_j$ will be made based on the index of the base classifier $j$ and the output vector of the classifier

$$\mathbf{f}^{(j)}(\mathbf{x}) = \sum_{j'=1}^{j} b_{j'}(\mathbf{x})\mathbf{h}_{j'}(\mathbf{x}). \quad (3)$$

built up to step $j$.[1] Formally, $a_j = \pi\big((\mathbf{s}_j(\mathbf{x}))\big)$, where

$$\mathbf{s}_j(\mathbf{x}) = \big(f_1^{(j-1)}(\mathbf{x}), \ldots, f_K^{(j-1)}(\mathbf{x}), j-1\big) \in \mathbb{R}^K \times \mathbb{N}^+ \quad (4)$$

is the *state* we are in before visiting $\mathbf{h}_j$, and $\pi$ is a *policy* that determines the action in state $\mathbf{s}_j$. The initial state $\mathbf{s}_1$ is the zero vector with $K + 1$ elements.

This setup formally defines a Markov decision process (MDP). An MDP is a 4-tuple $\mathcal{M} = (\mathcal{S}, \mathcal{A}, \mathcal{P}, \mathcal{R})$, where $\mathcal{S}$ is the (possibly infinite) state space and $\mathcal{A}$ is the countable set of actions. $\mathcal{P} : \mathcal{S} \times \mathcal{S} \times \mathcal{A} \to [0, 1]$ is the transition probability kernel which defines the random transitions $\mathbf{s}^{(t+1)} \sim \mathcal{P}(\cdot|\mathbf{s}^{(t)}, a^{(t)})$ from a state $\mathbf{s}^{(t)}$ applying the action $a^{(t)}$, and $\mathcal{R} : \mathbb{R} \times \mathcal{S} \times \mathcal{A} \to [0, 1]$ defines the distribution $\mathcal{R}(\cdot|\mathbf{s}^{(t)}, a^{(t)})$ of the *immediate reward* $r^{(t)}$ for each state-action pair. A *deterministic policy* $\pi$ assigns an action to each state $\pi : \mathcal{S} \to \mathcal{A}$. We will only use *undiscounted* and *episodic* MDPs where the policy $\pi$ is evaluated using the *expected sum of rewards*

$$\varrho = \mathbb{E}\left\{\sum_{t=1}^{T} r^{(t)}\right\} \quad (5)$$

with a finite horizon $T$. In the episodic setup we also have an *initial* state ($\mathbf{s}_1$ in our case) and a *terminal* state $\mathbf{s}_\infty$ which is impossible to leave. In our setup, the state $\mathbf{s}^{(t)}$ is equivalent to $\mathbf{s}_j(\mathbf{x})$ (4) with $j = t$. The action QUIT brings the process to the terminal state $\mathbf{s}_\infty$. Note that in $\mathbf{s}^{(T)}$ only the QUIT action is allowed.

---

[1] When using ADABOOST.MH, the base classifiers are binary $\mathbf{h}_j(\mathbf{x}) = \{\pm\alpha_j\}^K$, and we normalize the output (3) by $\sum_{j=1}^{N} \alpha_j$, but since this factor is constant, the only reason to do so is to make the range of the state space uniform across experiments.

## 2.1. The rewards

As our primary goal is to achieve a good performance in terms of the evaluation metric of interest, we will penalize the error of $\mathbf{f}^{(t)}$ when the action $a^{(t)} = $ QUIT is applied. The setup can handle any loss function. Here, we will use the *multi-class 0-1 loss function*

$$L_{\mathbb{I}}(\mathbf{f}, (\mathbf{x}, \ell)) = \mathbb{I}\left\{f_\ell(\mathbf{x}) - \max_{\ell' \neq \ell} f_{\ell'}(\mathbf{x}) < 0\right\}$$

and the *multi-class exponential loss function*

$$L_{\text{EXP}}(\mathbf{f}, (\mathbf{x}, \ell)) = \exp\left(\sum_{\ell' \neq \ell}^{K} f_{\ell'}(\mathbf{x}) - f_\ell(\mathbf{x})\right),$$

where the training observations $(\mathbf{x}, \ell) \in \mathbb{R}^d \times \{1, \ldots, K\}$ are drawn from a distribution $\mathfrak{D}$. Note that in the binary case, $L_{\mathbb{I}}$ and $L_{\text{EXP}}$ recover the classical binary notions.

With these notations, the reward for the QUIT action comes from the distribution

$$\mathcal{R}(r|\mathbf{s}^{(t)}, \text{QUIT}) = \\ \mathcal{P}_{(\mathbf{x}, \ell) \sim \mathfrak{D}}\big(-L(\mathbf{f}, (\mathbf{x}, \ell))|\mathbf{s}^{(t)} = (\mathbf{f}^{(t-1)}(\mathbf{x}), t-1)\big). \quad (6)$$

From now on we will refer to our algorithm as MDDAG.$\mathbb{I}$ or MDDAG.EXP when we use 0-1 loss or exponential loss, respectively. In principle, any of the usual convex upper bounds (e.g., logistic, hinge, quadratic) could be used in the MDP framework. The exponential loss function was inspired by the setup of ADABOOST (Freund & Schapire, 1997; Schapire & Singer, 1999).

To encourage sparsity, we will also penalize each evaluated base classifier $\mathbf{h}$ by a uniform fixed negative reward

$$\mathcal{R}(r|\mathbf{s}, \text{Eval}) = \delta(-\beta - r), \qquad (7)$$

where $\delta$ is the Dirac delta and $\beta$ is a hyperparameter that represents the accuracy-speed trade-off. Note that, again, this flexible setup can accommodate any cost function penalizing the evaluation of base classifiers. Finally, choosing the Skip action does not incur any reward, so $\mathcal{R}(r|\mathbf{s}, \text{Skip}) = \delta(0)$.

The goal of reinforcement learning (RL) in our case is to learn a policy which maximizes the expected sum of rewards (5). Since in our setup, the transition $\mathcal{P}$ is deterministic *given* the observation $\mathbf{x}$, the expectation in (5) is taken with respect to the random input point $(\mathbf{x}, \ell)$. This means that the global objective of the MDP is to minimize

$$\mathbb{E}_{(\mathbf{x}, \ell) \sim \mathfrak{D}} \left\{ L\big(\mathbf{f}, (\mathbf{x}, \ell)\big) + \beta \sum_{j=1}^{N} b_j(\mathbf{x}) \right\}. \qquad (8)$$

## 2.2. Learning the policy

There are several efficient algorithms available for learning the policy $\pi$ using an iid sample $\mathcal{D} = \big((\mathbf{x}_1, \ell_1), \ldots, (\mathbf{x}_n, \ell_n)\big)$ drawn from $\mathfrak{D}$ (Sutton & Barto, 1998). When $\mathcal{P}$ and $\mathcal{R}$ are unknown, model-free methods are commonly used for learning the policy $\pi$. These methods directly learn a *value function* (the expected reward in a state or for a state-action pair) and derive a policy from it. Among model-free RL algorithms, *temporal-difference* (TD) learning algorithms are the most widely used. They can be divided into two groups: *off-policy* and *on-policy* methods. In the case of off-policy methods the policy search method learns about one policy while following another, whereas in the on-policy case the policy search algorithm seeks to improve the current policy by maintaining sufficient exploration. On-policy methods have an appealing practical advantage: they usually converge faster to the optimal policy than off-policy methods.

We shall use the SARSA($\lambda$) algorithm (Rummery & Niranjan, 1994) with *replacing traces* to learn the policy $\pi$. For more details, we refer the reader to (Szepesvári, 2010). SARSA($\lambda$) is an on-policy method, so to make sure that all policies can be visited with nonzero probability, we use an $\epsilon$-greedy exploration strategy. To be precise, we apply SARSA in an episodic setup: we use a random training instance $\mathbf{x}$ from $\mathcal{D}$ per episode. The instance follows the current policy with probability $1 - \epsilon$ and chooses a random action with probability $\epsilon$. The instance observes the immediate rewards defined based on some loss function, or (7) after each action. The policy is updated during the episode according to SARSA($\lambda$).

In our experiments we used AdaBoost.MH[2] to obtain a pool of weak classifiers $\mathcal{H}$, and the RL Toolbox 2.0[3] for training the MDDAG. We ran AdaBoost.MH for $N = 1000$ iterations, and then trained SARSA($\lambda$) on the same training set. The hyperparameters of SARSA($\lambda$) were kept constant throughout the experiments. We set $\lambda$ to 0.95. In principle, the learning rate should decrease to 0, but we found that this setting forced the algorithm to converge too fast to suboptimal solutions. Instead we set the learning rate to a constant 0.2, we evaluated the current policy after every 10000 episodes, and we selected the best policy based on their performance also on the training set (overfitting the MDP was a non-issue). The exploration term $\epsilon$ was decreased gradually as $0.3 \times 1 / \lceil \frac{10000}{\tau} \rceil$, where $\tau$ is the number of training episodes. We trained SARSA($\lambda$) for $10^6$ episodes.

As a final remark, note that maximizing (8) over the data set $\mathcal{D}$ is equivalent to minimizing a margin-based loss with an $L_0$ constraint. If $r_1$ (6) is used as a reward, the loss is also non-convex, but minimizing a loss with an $L_0$ constraint is NP-hard even if the loss is convex (Davis et al., 1997). So, what we are aiming at is an MDP-based heuristic to solve an NP-hard problem, something that is not without precedent (Ejov et al., 2004). This equivalence implies that even though the algorithm would converge in the ideal case (with a decreasing learning rate), in principle, convergence can be exponentially slow in $n$. In practice, however, we had no problem finding good policies in reasonable training time.

## 3. Experiments

In Section 3.1 we first verify the sparsity and heterogeneity hypotheses on a synthetic toy example. In Section 3.2, we compare MDDAG with state-of-the-art cascade detectors on three object detection benchmarks. After, in Section 3.3 we show how the multi-class version of MDDAG performs on a benchmark web page ranking problem.

## 3.1. Synthetic data

The aim of this experiment was to verify whether MDDAG can learn the subset of "useful" base classifiers in a *data-dependent* way. We created a two-dimensional binary dataset with real-valued features

---

where the positive class was composed of two easily separable clusters (see Figure 1(a)). This is a typical case where ADABOOST or a traditional cascade is suboptimal since they both have to use *all* the base classifiers for all the positive instances (Bourdev & Brandt, 2005).

We ran MDDAG.I with $\beta = 0.01$ on the 1000 decision stumps learned by ADABOOST.MH. In Figure 1(b), we plot the number of base classifiers used for each individual positive instance as a function of the two-dimensional instance itself. As expected, the "easier" the instance, the smaller the number of base classifiers are needed for classification. Figure 1(c) confirms our second hypothesis: base classifiers are used selectively, depending on whether the positive instance is in the blue or red cluster.

The lower panel of Figure 1 shows a graphical representation of the MDDAG classifier **f** acting on a data set $\mathcal{D}$. The nodes of the directed acyclic graph (DAG) are the base classifiers in $\mathcal{H}$. Each observation $(\mathbf{x}, \ell) \in \mathcal{D}$ determines a set of edges

$$U_{\mathbf{x}} = \{(j, j') : b_j(\mathbf{x}) = b_{j'}(\mathbf{x}) = 1 \wedge$$
$$b_{j''}(\mathbf{x}) = 0 \text{ for all } j < j'' < j'\}.$$

In other words, we take all the base classifiers that are evaluated on the instance $(\mathbf{x}, \ell)$ and connect the nodes representing these base classifiers with a direct edge. The edge set $U_{\mathbf{x}}$ is called the *classification path* of $\mathbf{x}$ which constitutes a directed path by definition. The DAG we plot in Figure 1(d) includes all of the edges $U = \bigcup_{(\mathbf{x}, 1) \in \mathcal{D}} U_{\mathbf{x}}$ generated by the *positive* instances taken from the training data $\mathcal{D}$. The width of an edge $(j, j')$ is proportional to its *multiplicity* $\#\{\mathbf{x} : (j, j') \in U_{\mathbf{x}}, (\mathbf{x}, 1) \in \mathcal{D}\}$. The color of an edge $(j, j')$ represents the proportion of observations taken from the blue and red sub-classes, whose classification path includes $(j, j')$. Similarly, the size of the node is proportional to $\#\{\mathbf{x} : b_j(\mathbf{x}) = 1, (\mathbf{x}, 1) \in \mathcal{D}\}$, and the color of the nodes represent sub-class proportions. The structure of the DAG also agrees with our original intuition, namely that the bulk of the two sub-classes are separated early and follow different classification paths. It is also worth noting that even though the number of possible classification paths is exponentially large, the number of realized paths is quite small. Some "noisy" points along the main diagonal (border between the subclasses) generate rare subpaths, but the bulk of the data mostly follows two paths.

### 3.2. Binary detection benchmarks

In these experiments we applied MDDAG on three image data sets often used for benchmarking object detection cascades. VJ (Viola & Jones, 2004) and CBCL are face recognition benchmarks, and DPED (Munder & Gavrila, 2006) is a pedestrian recognition data set. We divided the data sets into training and test sets.

We compared MDDAG to three state-of-the-art object detection algorithms (the original Viola-Jones cascade VJCASCADE (Viola & Jones, 2004), FC-BOOST (Saberian & Vasconcelos, 2010), and SOFT-CASCADE (Bourdev & Brandt, 2005)). VJCASCADE builds the cascade stage-by-stage by running ADABOOST in each stage. It stops adding base classifiers to the $m$th stage when the false positive rate (FPR) falls below $p_{\text{FPR}}^m$ and true positive rate (TPR) exceeds $p_{\text{TPR}}^m$, where $p_{\text{TPR}}$ and $p_{\text{FTR}}$ are hyperparameters of the algorithm. The total number of stages is also a hyperparameter. FCBOOST also adds base classifiers iteratively to the cascade, but the base classifier can be inserted into any of the stages. The goal is to minimize a global criterion which, similarly to (8), is composed of a performance-based term and a complexity-based term. The number of iterations and the parameter $\eta$ that determines the trade-off between the two competing objectives are hyperparameters of the algorithm. SOFTCASCADE, like MDDAG, builds a cascade on the output of ADABOOST, where each stage consists of exactly one base classifier. The final number of base classifiers is decided beforehand by the hyperparameter $N$. In each iteration $j$, the base classifier with the highest balanced edge is selected, and the detection threshold $\theta_j$ is set to achieve a TPR of $1 - \exp(\alpha j / N - \alpha \mathbb{I}\{\alpha < 0\})$, where $\alpha$ is a second hyperparameter of the algorithm. Both the TPR and the number of base classifiers increase with $\alpha$, so the choice of $\alpha$ influences the speed/accuracy trade-off (although not as explicitly as our $\beta$ or FCBOOST's $\eta$).

Comparing test-time-constrained detection algorithms is quite difficult. The usual trade-off between the false positive rate (FPR) and the true positive rate (TPR) can be captured by ROC curves, but here we also have to take into account the computational efficiency of the detector. In (Bourdev & Brandt, 2005) this problem is solved by displaying three-dimensional FPR/TPR/number-of-features surfaces. Here, we decided to show two-dimensional slices of these surfaces: we fix the FPR to reasonable values and plot the TPR against the detection time. In each of the algorithms we used Haar features as base classifiers, so the detection time can be uniformly measured in terms of the average number of base classifiers needed for detection. In typical detection problems the number of background (negative) instances is orders of magnitudes higher than the number of signal (positive) instances, so we computed this average only over the negative test set. It turns out that using this measure, vanilla ADABOOST is fairly competitive with tailor-made cascade detectors, so we also included it in the comparison.

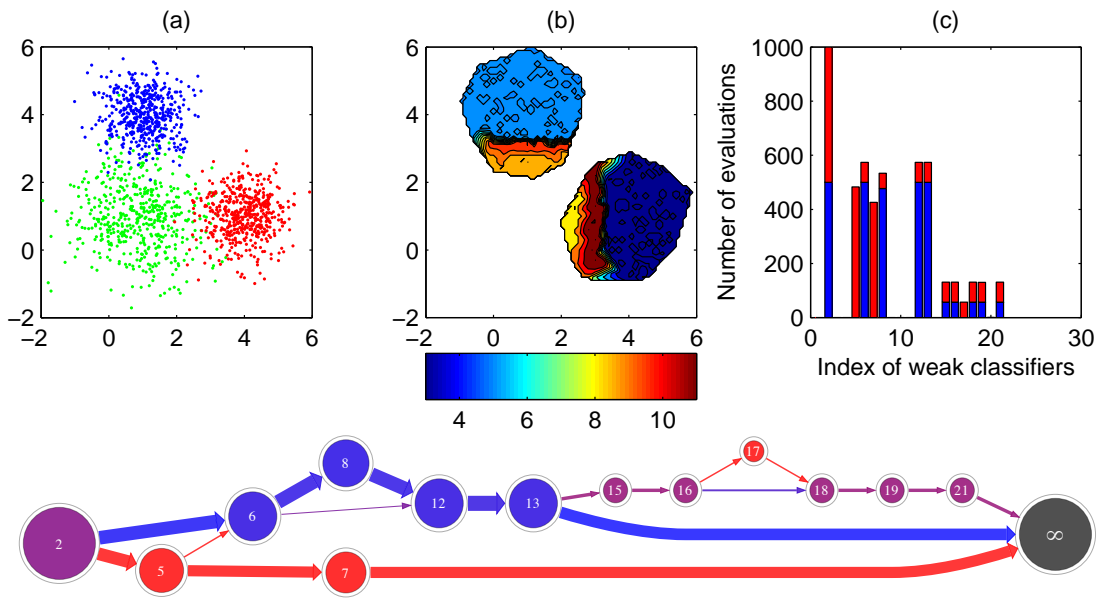

*Figure 1.* Experiments with synthetic data. (a) The positive class is composed of the blue and red clusters, and the negative class is the green cluster. (b) The number of base classifiers used for each individual positive instance versus the two-dimensional feature coordinates. (c) The number of positive instances from the blue/red clusters on which a given base classifier was applied according to the policy learned by MDDAG.I. Lower panel: the decision DAG for the positive class.

Computing the TPR versus number-of-features curve at a fixed FPR cannot be done in a generic algorithm-independent way. For ADABOOST, in each iteration $j$ (that is, for each number $j$ of base classifiers) we tune the detection threshold $\theta$ to achieve the given test FPR, and plot the achieved test TPR versus $j$. In the other three algorithms we have 2-3 hyperparameters that explicitly or implicitly influence the average number of base classifiers and the overall performance. We ran the algorithms using different hyperparameter combinations. In each run we set the detection threshold $\theta$ to achieve the given test FPR. With this threshold, each run $k$ determines a TPR/average-number-of-features pair $(p_k, N_k)$ on the training set and $(p'_k, N'_k)$ on the test set. For each $N$, we find the run $k^*(N) = \arg \max_{k : N_k \leq N} p_k$ that achieves the best *training* $p_k$ using at most $N$ base classifiers, and plot the *test* TPR $p'_{k^*(N)}$ versus $N$. Although overfitting is not an issue here (the complexity of the classifiers is relatively low), this setup is important for a fair comparison. If overfitting were an issue, the optimization could also be carried out on a validation set, independent of both the test and the training sets. Optimizing the TPR on the training set and plotting it on the test set also explains why the curves are non-monotonic.

Figure 2 shows the results we obtained. Although the differences are quite small, MDDAG outperforms the three benchmarks algorithms consistently in the regime of low number base classifiers, and it is competitive with them over the full range. MDDAG.I is slightly better at low complexities, whereas MDDAG.EXP is more effective at a slightly higher number of base classifiers. This is not surprising as in the low complexity regime the 0-1 error is more aggressive and closer to the measured TPR, whereas when most of the instances are classified correctly, without the margin information it is impossible to improve the policy any further.

## 3.3. Ranking with multi-class DAGs

Although object detection is arguably the best-known test-time-constrained problem, it is far from being unique. In web page ranking, the problem is similar: training time can be almost unlimited, but the learned ranker must be fast to execute. State-of-the-art techniques often use thousands of trained models in an ensemble setup (Chapelle et al., 2011b), so extracting lean rankers from the full models is an important practical issue. One of the difficulties in this case is that relevance labels may be non-binary, so classical object-detection cascades cannot be applied. At the same time, the principles used to design cascades re-surface also in this domain, although the setup is rather new and the algorithms require a fair amount

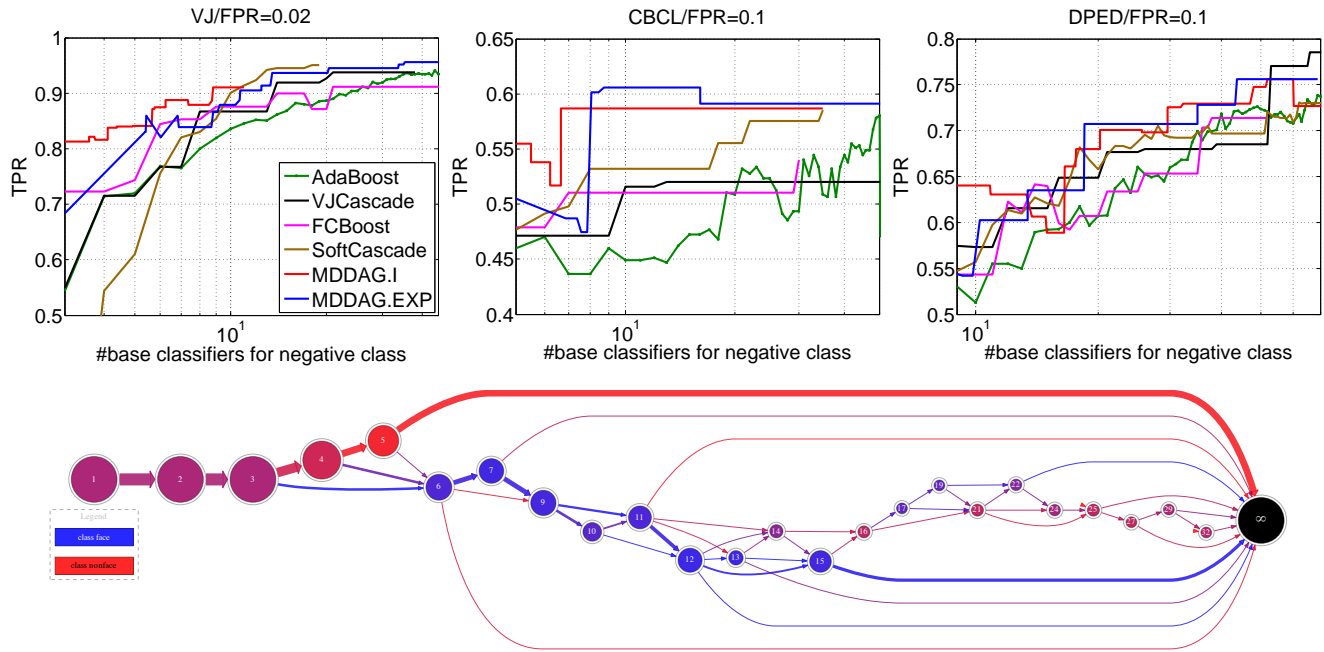

*Figure 2.* The true positive rate (TPR) vs. the average number of base classifiers evaluated on negative test instances. The results are computed for different fixed false positive rates (FPR) shown in the title of panels. Lower panel: one of the DAGs learned for the VJ database.

of manual tuning (Cambazoglu et al., 2010). Despite this, MDDAG can be used for this task *as is*.

To evaluate MDDAG on a multi-class classification/ranking problem, we present results on the MQ2007 and MQ2008 data sets taken from LETOR 4.0. In web page ranking, observations come in the form of query-document pairs, and the performance of the ranker is evaluated using tailor-made loss or gain functions that take as input the ordering of all the documents given a query. To train a ranker, query-document pairs come with manually annotated *relevance labels* that are usually multi-valued ($\{0, 1, 2\}$ in our case). One common performance measure is the Normalized Discounted Cumulative Gain (NDCG$_m$) (Järvelin & Kekäläinen, 2002) which is based on the first $m$ documents in the order output by the ranker. We used the *averaged* NDCG score $\overline{\text{NDCG}}$, provided by LETOR 4.0, that takes an average of the query-wise NDCG$_m$ values to evaluate the algorithms. In these experiments the base learners were decision trees with eight leaves.

The goal of MDDAG is similar to the binary case, namely to achieve a comparable performance to AdaBoost.MH using fewer base learners. To make the comparison fair, we employed the same calibration method to convert the output of the multi-class clas-sifiers to a scoring function and then to a ranking (Li et al., 2007). Since the goal this time was not detection, we simply evaluated the average NDCG for each run $k$ to obtain $(\overline{\text{NDCG}}_k, N_k)$ on the training set and $(\overline{\text{NDCG}}'_k, N'_k)$ on the test set. We then selected $k^*(N) = \arg \max_{k: N_k \leq N} \overline{\text{NDCG}}_k$ and plotted the test $\text{NDCG}'_{k^*(N)}$ against $N$. Figure 3 tells us that MDDAG performs as well as AdaBoost.MH with roughly two-fold savings in the number of base classifiers.

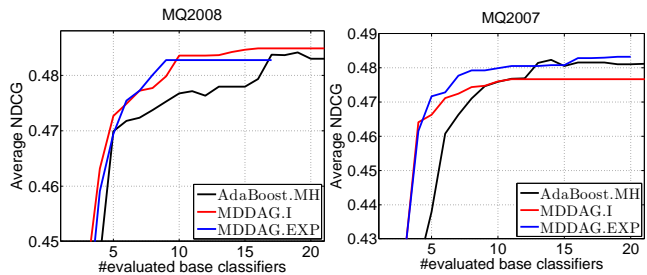

*Figure 3.* The average NDCG vs. the average number of base classifiers evaluated on test queries.

## 4. Related works

Besides (Póczos et al., 2009) and (Dulac-Arnold et al., 2011), MDDAG has several close relatives in the family of supervised methods. It is obviously related to algorithms taken from the vast array of sparse meth-ods. The main advantage here is that the MDP setup allows one to achieve sparsity in a dynamical data-dependent way. This feature relates the technique to unsupervised sparse *coding* (Lee et al., 2007; Ranzato et al., 2007) rather than to sparse classification or regression. On a more abstract level, MDDAG is also similar to (Larochelle & Hinton, 2010)'s approach to "learn where to look". Their goal is to find a sequence of two-dimensional features for classifying images in a data-dependent way, whereas we do a similar search in a one-dimensional ordered sequence of features.

## 5. Conclusions

In this paper, we introduced an MDP-based design of decision DAGs. The output of the algorithm is a data-dependent sparse classifier, which means that every instance "chooses" the base classifiers or features that it needs to predict its class index. The algorithm is competitive with state-of-the-art cascade detectors on object detection benchmarks, and it is also directly applicable to test-time-constrained problems involving multi-class classification, such as web page ranking. However, in our view, the main benefit of the algorithm is not necessarily its performance, but its simplicity and versatility. First, MDDAG is basically a turn-key procedure: it comes with one user-provided hyperparameter with a clear semantics of directly determining the accuracy-speed trade-off. Second, MDDAG can be readily applied to problems different from classification by redefining the rewards on the QUIT and EVAL actions. For example, one can easily design *regression* or *cost-sensitive* classification DAGs by using an appropriate reward in (6), or add a weighting to (7) if the features have different evaluation costs.

## Acknowledgements

This work was supported by the ANR-2010-COSI-002 grant of the French National Research Agency.

## References

Benbouzid, D., Busa-Fekete, R., Casagrande, N., Collin, F.-D., and Kégl, B. MultiBoost: a multi-purpose boosting package. *JMLR*, 13:549–553, 2012.

Bourdev, L. and Brandt, J. Robust object detection via soft cascade. In *CVPR*, volume 2, pp. 236–243, 2005.

Cambazoglu et al. Early exit optimizations for additive machine learned ranking systems. In *WSDM*, pp. 411–420, 2010.

Chapelle, O., Chang, Y., and Liu, T.Y. Future directions in learning to rank. In *JMLR W&CP*, volume 14, pp. 91–100, 2011a.

Chapelle, O., Chang, Y., and Liu, T.Y. (eds.). *Ya-hoo! Learning-to-Rank Challenge*, volume 14 of *JMLR W&CP*, 2011b.

Chapelle, Olivier and Chang, Yi. Yahoo! Learning-to-Rank Challenge overview. In *JMLR W&CP*, volume 14, pp. 1–24, 2011.

Davis, G., Mallat, S., and Avellaneda, M. Adaptive greedy approximations. *Constructive Approximation*, 13(1):57–98, 1997.

Dulac-Arnold, G., Denoyer, L., Preux, P., and Gallinari, P. Datum-wise classification: A sequential approach to sparsity. In *ECML*, 2011.

Ejov, V., Filar, J., and Gondzio, J. An interior point heuristic for the Hamiltonian cycle problem via Markov Decision Processes. *JGO*, 29(3):315–334, 2004.

Freund, Y. and Schapire, R. E. A decision-theoretic generalization of on-line learning and an application to boosting. *JCSS*, 55:119–139, 1997.

Gligorov, V. A single track HLT1 trigger. Technical Report LHCb-PUB-2011-003, CERN, 2011.

Järvelin, K. and Kekäläinen, J. Cumulated gain-based evaluation of IR techniques. *ACM TIS*, 20:422–446, 2002.

Larochelle, H. and Hinton, G. Learning to combine foveal glimpses with a third-order Boltzmann machine. In *NIPS*, pp. 1243–1251, 2010.

Lee, H., Battle, A., Raina, R., and Ng, A. Y. Efficient sparse coding algorithms. In *NIPS*, pp. 801–808, 2007.

Li, P., Burges, C., and Wu, Q. McRank: Learning to rank using multiple classification and gradient boosting. In *NIPS*, pp. 897–904, 2007.

Munder, S. and Gavrila, D. M. An experimental study on pedestrian classification. *IEEE PAMI*, 28:1863–1868, 2006.

Póczos, B., Abbasi-Yadkori, Y., Szepesvári, Cs., Greiner, R., and Sturtevant, N. Learning when to stop thinking and do something! In *ICML*, pp. 825–832, 2009.

Ranzato, M., Poultney, C., Chopra, S., and LeCun, Y. Efficient learning of sparse representations with an energy-based model. In *NIPS*, pp. 1137–1144, 2007.

Rummery, G. A. and Niranjan, M. On-line Q-learning using connectionist systems. Technical Report CUED/F-INFENG/TR 166, Cambridge University, 1994.

Saberian, M. and Vasconcelos, N. Boosting classifier cascades. In *NIPS*, pp. 2047–2055, 2010.

Schapire, R.E. and Singer, Y. Improved boosting algorithms using confidence-rated predictions. *Machine Learning*, 37(3):297–336, 1999.

Sutton, R.S. and Barto, A.G. *Reinforcement learning: an introduction*. Adaptive computation and machine learning. MIT Press, 1998.

Szepesvári, Cs. *Algorithms for Reinforcement Learning*. Morgan and Claypool, 2010.

Viola, P. and Jones, M. Robust real-time face detection. *IJCV*, 57:137–154, 2004.